\documentclass[runningheads]{llncs}
 
\usepackage{eccv}

\usepackage{eccvabbrv}

\usepackage{graphicx}
\usepackage{booktabs}

\usepackage[dvipsnames]{xcolor}
\usepackage{multirow} 
\usepackage[accsupp]{axessibility}  

\usepackage[pagebackref,breaklinks,colorlinks,citecolor=eccvblue]{hyperref}

\usepackage{orcidlink}

\begin{document}
\title{Towards Video Anomaly Detection from Event Streams: A Baseline and Benchmark Datasets} 
\titlerunning{EWAD}


\author{Peng Wu\inst{1} \and Yuting Yan\inst{1} \and Guansong Pang\inst{2} \and Yujia Sun\inst{3} \and Qingsen Yan\inst{1} \and Peng Wang\inst{1} \and Yanning Zhang\inst{1}}

\authorrunning{P. Wu et al.}

\institute{School of Computer Science, Northwestern Polytechnical University, China \\
\email{\{xdwupeng, xgdyanyuting, qingsenyan\}@gmail.com, \{peng.wang, ynzhang\}@nwpu.edu.cn} \and
School of Computing and Information Systems, Singapore Management University, Singapore \\
\email{gspang@smu.edu.sg} \and
School of Artificial Intelligence, Xidian University, China \\
\email{yjsun@stu.xidian.edu.cn}}

\maketitle

\begin{abstract}
Event-based vision, characterized by low redundancy, focus on dynamic motion, and inherent privacy-preserving properties, naturally fits the demands of video anomaly detection (VAD). However, the absence of dedicated event-stream anomaly detection datasets and effective modeling strategies has significantly hindered progress in this field. 
In this work, we take the first major step toward establishing event-based VAD as a unified research direction. 
{{We first construct multiple simulated event-stream based benchmarks for}} video anomaly detection, featuring synchronized event and RGB recordings. Leveraging the unique properties of events, we then propose an EVent-centric spatiotemporal Video Anomaly Detection framework, namely \textbf{EWAD}, with three key innovations: an event density aware dynamic sampling strategy to select temporally informative segments; a density-modulated temporal modeling approach that captures contextual relations from sparse event streams; and an RGB-to-event knowledge distillation mechanism to enhance event-based representations under weak supervision. Extensive experiments on three benchmarks demonstrate that our EWAD achieves significant improvements over existing approaches, highlighting the potential and effectiveness of event-driven modeling for video anomaly detection. 
The benchmark datasets and code are available at \url{https://github.com/kanyutingfeng/EWAD}.
  \keywords{Video anomaly detection \and Event-based vision \and Vision-language model}
\end{abstract}

\section{Introduction}
With the increasing demand for video understanding in public safety, video anomaly detection (VAD) has emerged as a critical task for identifying potential risks and abnormal behaviors~\cite{liu2023generalized, barbosa2025survey}. In recent years, driven by advances in deep learning, RGB-based VAD methods have achieved remarkable progress. However, RGB videos inherently suffer from fixed frame rates, high data redundancy, and delayed dynamic perception, limiting their effectiveness in high-density or rapidly changing environments. 

In contrast, event cameras, as an emerging type of vision sensor, offer an alternative paradigm for video analysis. Operating asynchronously at the pixel level, they emit events only when brightness changes exceed a threshold, encoding the time, location, and polarity of such changes. This results in high temporal resolution, low latency and inherently sparse outputs~\cite{liu2025eventgpt}. Rather than densely capturing static scenes, event cameras naturally emphasize motion and dynamic changes, attributes highly aligned with the nature of anomalies in videos. These characteristics make event data particularly well-suited for the VAD task, as they enhance temporal sensitivity, reduce redundant background noise, and preserve privacy.

Despite growing interest in event-based vision for low-level tasks such as image restoration, its application to VAD remains largely unexplored. Two key challenges hinder progress in this field: the lack of high-quality event-stream VAD datasets~\cite{qian2025ucf}, and the limited transferability of existing VAD models originally designed for synchronous RGB videos to the asynchronous and sparse nature of event data. Currently, there has been little research on effective modeling strategies capable of fully leveraging the spatiotemporal representation inherent in event streams. For example, Qian et al.~\cite{qian2025ucf} proposed a VAD framework based on the spiking neural network (SNN) using event data, demonstrating preliminary effectiveness. However, this line of research largely overlooks the powerful cross-modal alignment capabilities of recent vision-language models (VLMs) and the complementary information available in paired RGB frames. 

In this paper, we propose a unified framework that tackles two key challenges: the scarcity of dedicated benchmarks and the difficulty of modeling event-based spatiotemporal dynamics. 
{{We first construct large-scale simulated event-based benchmarks}} through temporal alignment of event streams with widely used RGB-based VAD datasets (UCF-Crime~\cite{sultani2018real}, CCTV-Fight~\cite{perez2019detection}, and UBnormal~\cite{acsintoae2022ubnormal}). These benchmarks enable systematic evaluation of event-based anomaly detection across diverse scenes and scenarios. 
Building upon this foundation, we then develop an \textbf{EV}ent-centric \textbf{V}ideo \textbf{A}nomaly \textbf{D}etection baseline named \textbf{EWAD} tailored to spatiotemporal event understanding. 
Unlike conventional cameras that capture RGB frames at fixed intervals, event cameras asynchronously record changes in brightness at the pixel level, producing sparse signals that primarily reflect scene dynamics. Such representations tend to highlight motion-related patterns while suppressing static background information. These characteristics provide complementary cues to frame-based representations and may be beneficial for motion-centric tasks such as anomaly detection. Motivated by this observation, EWAD is designed to better exploit the temporal dynamics and sparse structure of event streams for anomaly detection.

Specifically, we first propose an event-density aware dynamic sampling (EDS) strategy that adaptively selects temporally diverse and discriminative frames, i.e., emphasizing those moments likely to contain anomalous cues, thus effectively leveraging the inherently sparse and bursty nature of event data while ensuring comprehensive temporal coverage critical for efficent anomaly detection. Then, we design an event-modulated distance-decay attention (EDA) mechanism that jointly encodes event density and inter-event intervals, allowing the model to capture long-range temporal dependencies by dynamically adjusting time perception. By aligning attention dynamics with event saliency, this mechanism helps preserve sparsity while enhancing temporal awareness. Notably, this mechanism is lightweight and can be seamlessly integrated into existing temporal modeling modules. Furthermore, to compensate for the limited supervisions provided by hard labels and deficient information inherent in event-only data, we propose a cross-modal knowledge distillation (KD) strategy, wherein high-level priors from pretrained RGB models are transferred to the event-based model. This distillation strategy enables the event model (student) to acquire the implicit knowledge embedded in the teacher model, such as inter-class similarity relationships and the structural organization of the feature space. 
This distillation mechanism is only applied during training stage. 
Finally, we extend our method to explore event-based spatial anomaly localization, demonstrating the capability of event streams in delivering fine-grained cues for regional abnormality understanding.

In summary, our contributions are four-fold:
\begin{itemize}
\item We introduce large-scale benchmarks that align event streams with RGB datasets to enable systematic evaluation of event-centric video anomaly detection across varied scenes.
\item We propose a dynamic sampling strategy and an event-modulated attention mechanism tailored to event streams, improving temporal sensitivity and efficiency.
\item We develop a knowledge distillation framework that transfers high-level semantics from RGB-based models to enhance the representational capacity of event-only detectors.
\item Extensive experiments on three benchmarks demonstrate that our EWAD achieves state-of-the-art performance among event-based methods and establishes a strong baseline for future research in event-centric video anomaly detection.
\end{itemize}

\section{Related Work}
\subsection{Deep Learning for Video Anomaly Detection}
Deep learning methods have significantly advanced VAD by enabling data-driven deep neural network (DNN) pipelines. However, their success often hinges on the availability of large-scale, finely annotated datasets, which are expensive and time-consuming to obtain. To address this limitation, weakly supervised learning has emerged as a promising alternative. It reduces the annotation burden while maintaining competitive detection performance, and has become a vibrant research direction in recent years. Most weakly supervised methods adopt a multiple instance learning (MIL) framework~\cite{huang2022weakly}, where only video-level labels are provided, and the goal is to infer segment- or frame-level anomaly scores. A seminal work by Sultani et al.~\cite{sultani2018real} introduced MIL to VAD, laying the foundation for subsequent weakly supervised approaches. 
He et al.~\cite{he2024adversarial} tackled the imbalance between normal and abnormal videos via adversarial training and hard-negative mining to enhance detection with limited anomalies. Su et al.~\cite{su2025semantic} proposed a semantic-driven consistency scheme to align object localization and features, enabling interpretable anomaly detection.

With the rapid development of VLMs, especially the CLIP model \cite{radford2021learning}, cross-modal semantic understanding has been significantly enhanced. CLIP projects both images and textual descriptions into a shared embedding space, enabling efficient visual-textual alignment and demonstrating impressive performance across a wide range of vision tasks~\cite{wu2025varcmp}. In the context of weakly supervised VAD, a number of recent methods, such as VadCLIP~\cite{wu2024vadclip}, OVVAD~\cite{wu2024open}, STPrompts~\cite{wu2024weakly}, and TPWNG~\cite{yang2024text}, leverage CLIP’s vision-language alignment by incorporating textual prompts, which embed high-level prior knowledge into the anomaly detection process. 
Building on the powerful reasoning capabilities of VLMS and large language models (LLMs)~\cite{tang2024hawk}, several inference-based approaches have emerged recently, including LAVAD~\cite{zanella2024harnessing}, Holmes-VAU~\cite{zhang2025holmes}, Vera~\cite{ye2025vera}, Ex-VAD~\cite{huangex}, and SlowfastVAD~\cite{ding2025slowfastvad}. These methods eliminate the need for additional training and rely instead on zero-shot or few-shot inference mechanisms. While promising, such approaches often suffer from slower inference speed and have room for improvement in detection performance.

\subsection{Event Cameras in Computer Vision}
Event cameras are bio-inspired vision sensors that offer several unique advantages, including ultra-high temporal resolution, low latency, high dynamic range. Unlike conventional frame-based cameras, event cameras asynchronously capture per-pixel intensity changes, making them particularly suitable for high-speed motion analysis and low-light conditions. These properties have spurred interest in applying event cameras to a variety of computer vision tasks, including image restoration~\cite{wang2020event}, optical flow estimation~\cite{gehrig2024dense}, and object recognition~\cite{chen2025event}. 
Despite their advantages, the application of event cameras to video anomaly detection remains in its early stages. Event streams naturally suppress static background and emphasize dynamic changes, which can be advantageous for detecting anomalous behaviors. For instance, Qian et al.~\cite{qian2025ucf} extended the existing UCF-Crime dataset by collecting corresponding event data using an event camera, and proposed a SNN–based VAD framework that demonstrated preliminary effectiveness. Nonetheless, this line of research faces several limitations: the high acquisition cost associated with event cameras, the lack of diverse event-based datasets, the underutilization of the powerful multimodal capabilities of models like CLIP, the insufficient exploration of complementary information between RGB and event modalities, and the lack of investigation into the potential of event data’s low-redundancy characteristics for spatial localization.

\section{Dataset}
\subsection{Dataset Overview}
The quality of datasets fundamentally determines the performance and comparability of VAD models. However, the field currently faces a critical shortage of publicly available event-based VAD datasets, severely hindering research progress. Real event data collection is costly, highly sensitive to environmental conditions, and further complicated by the inherent rarity of anomalous events, making the construction of large-scale, real-world datasets exceedingly difficult.

To address this limitation, we adopt a simulation-based strategy to generate high-quality event data from widely used RGB-based VAD benchmarks, including UCF-Crime~\cite{sultani2018real}, CCTV-Fights~\cite{perez2019detection}, and UBNormal~\cite{acsintoae2022ubnormal}, {which cover diverse scenes and anomaly types.} Table~\ref{tab:datasets} summarizes detailed statistics of datasets. 

\begin{table*}[t]
    \centering
    \caption{Summary of event-based VAD datasets.}
    \begin{tabular}{lccccccc}
        \toprule
        Dataset & Total videos & Length & Train videos & Test videos & Classes & Scenes & Resolution \\
        \midrule
        UCF-Crime & 1900 & 128 hours & 1610 & 290 & 14 & - & 346 $\times$ 260 \\
        CCTV-Fights & 1000 & 18 hours & 750 & 250 & 1 & - & 346 $\times$ 260 \\
        UBnormal & 543 &  2.2 hours &268 & 211 & 23 & 29 & 640 $\times$ 480 \\
        \bottomrule
    \end{tabular}
    \label{tab:datasets}
\end{table*}

To the best of our knowledge, this work introduces the first publicly available large-scale event-based VAD benchmark constructed and diverse real-world scenarios, such as urban streets and indoor surveillance. The datasets cover anomalies including shooting, fall, fight, road accident, etc, with each video comprising millions of event points, highlighting the ultra-high temporal resolution of event cameras. This large-scale, high-fidelity resource provides a solid foundation for training and evaluating weakly supervised event-centric VAD models and is expected to significantly accelerate research in this emerging domain.

\subsection{Event Data Generation}
We employ the state-of-the-art v2e simulator~\cite{hu2021v2e}, which models optical flow and pixel-level brightness changes to convert RGB videos into realistic event streams.
{The use of v2e is motivated by its widespread adoption in the event-vision community, where multiple benchmarks and prior studies rely on simulated event data ~\cite{liu2023sensing,liu2026eventflash,li2022asynchronous,courtois2025spiking} for large-scale evaluation.}
This process preserves the spatiotemporal dynamics and anomaly labels of the original videos, ensuring temporal continuity and spatial consistency in the generated data. We upsampled all videos to 100 fps during simulation to leverage event cameras' high temporal resolution. 
{More information on setting the v2e simulator parameters can be found in the supplementary materials.}
As shown in Figure~\ref{fig-data}, the simulated event data clearly captures the motion of foreground objects while exhibiting minimal redundant dynamic visual information in the background.

\begin{figure*}[t]
    \centering
    \includegraphics[width=0.99\linewidth]{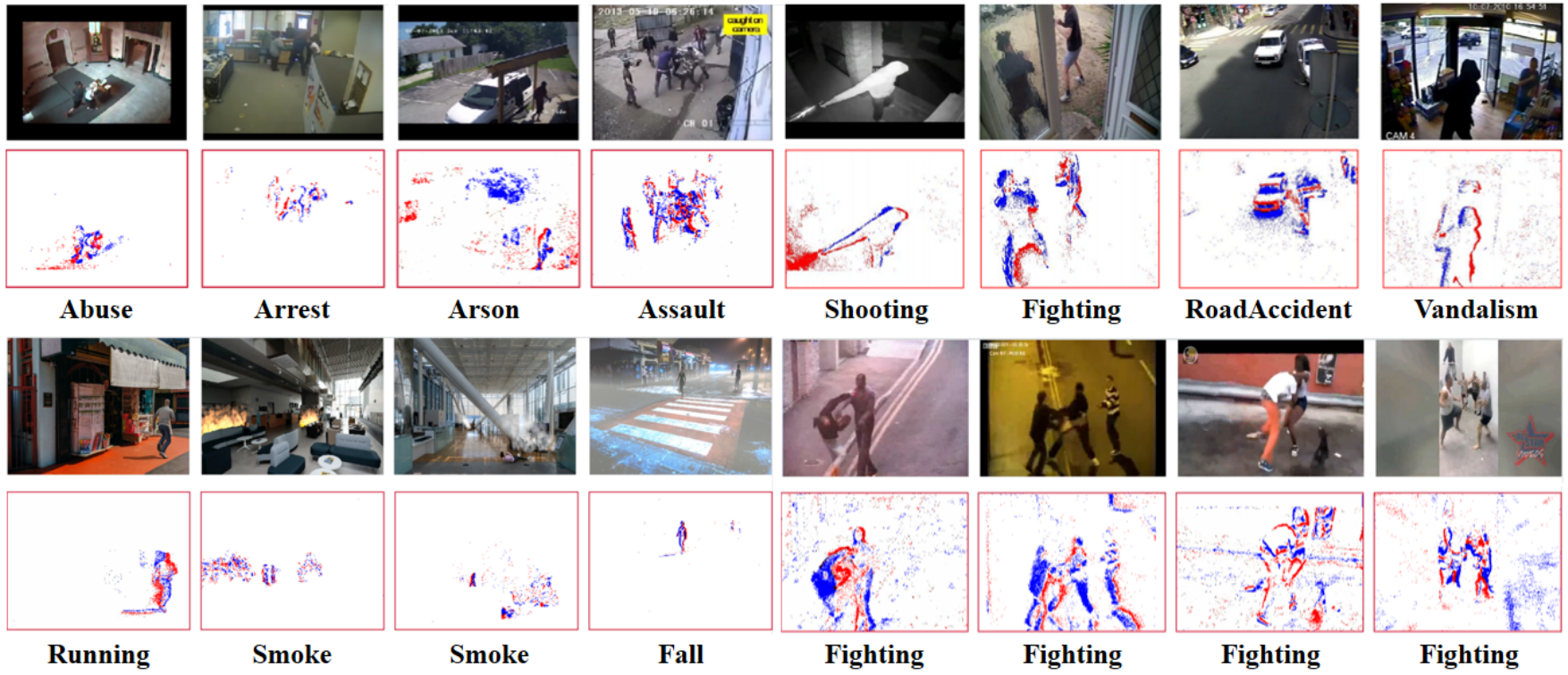}
    \caption{Examples of generated event data and RGB counterparts, sourced from UCF-Crime, UBnormal, and CCTV-Fight.} 
    \label{fig-data}
\end{figure*}

\subsection{Adaptive Event Frame Generation}
Event data can be represented in various forms. In this work, we adopt event frames by discretizing asynchronous events into frame sequences for compatibility with vision models. Traditional methods often fix the number of events per frame, which may cause sparsity-induced blur or density-induced ghosting. To mitigate this, we propose an adaptive frame generation strategy that adjusts event density, striking a balance between temporal resolution and data sparsity.

The size and timestamp of each bin are determined by following the feature extraction strategy used for RGB frames~\cite{sultani2018real}. 
To maintain consistency with the original RGB video's frame rate, we sample one frame every 16 frames as the center of a time window, and extend the window by 8 frames before and after the center, thereby dividing the event data into non-overlapping time segments to ensure each event frame corresponds to unique and non-redundant event data.

We set a baseline of one-tenth the average event points per bin (i.e., $\textit{mean} \times {1}/{10}$). 
We observe that frames with excessive events require downsampling to avoid ghosting artifacts, while those with insufficient events benefit from upsampling to preserve structural integrity. {Ghosting refers to density accumulation artifacts, not optical blur.} Given that the event count distribution is often skewed by a small number of high-activity frames, we empirically use the median rather than the mean event count per video as a more robust reference for sparsity control.
We define a sparsity coefficient as:
$sc = N_c $/$ median$,
where ${N_c}$ represents the number of event points contained within the bin, and a larger $sc$ indicates a denser frame and thus necessitates fewer events per bin, whereas a smaller $sc$ corresponds to a sparser frame requiring more events. The final number of event points per bin is computed as,
\begin{equation}
\textit{EventNum} = \frac{1}{10}\times \textit{mean}  + \frac{1}{sc} \times \textit{median}
\end{equation}
To further mitigate ghosting in videos characterized by intense global motion, we also enforce an upper bound $10000$ on the number of events per bin. This constraint ensures temporal coherence and avoids frame saturation in highly dynamic scenes.

\section{Methodology}
\subsection{Overview}
\begin{figure*}[t]
    \centering
    \includegraphics[width=0.99\linewidth]{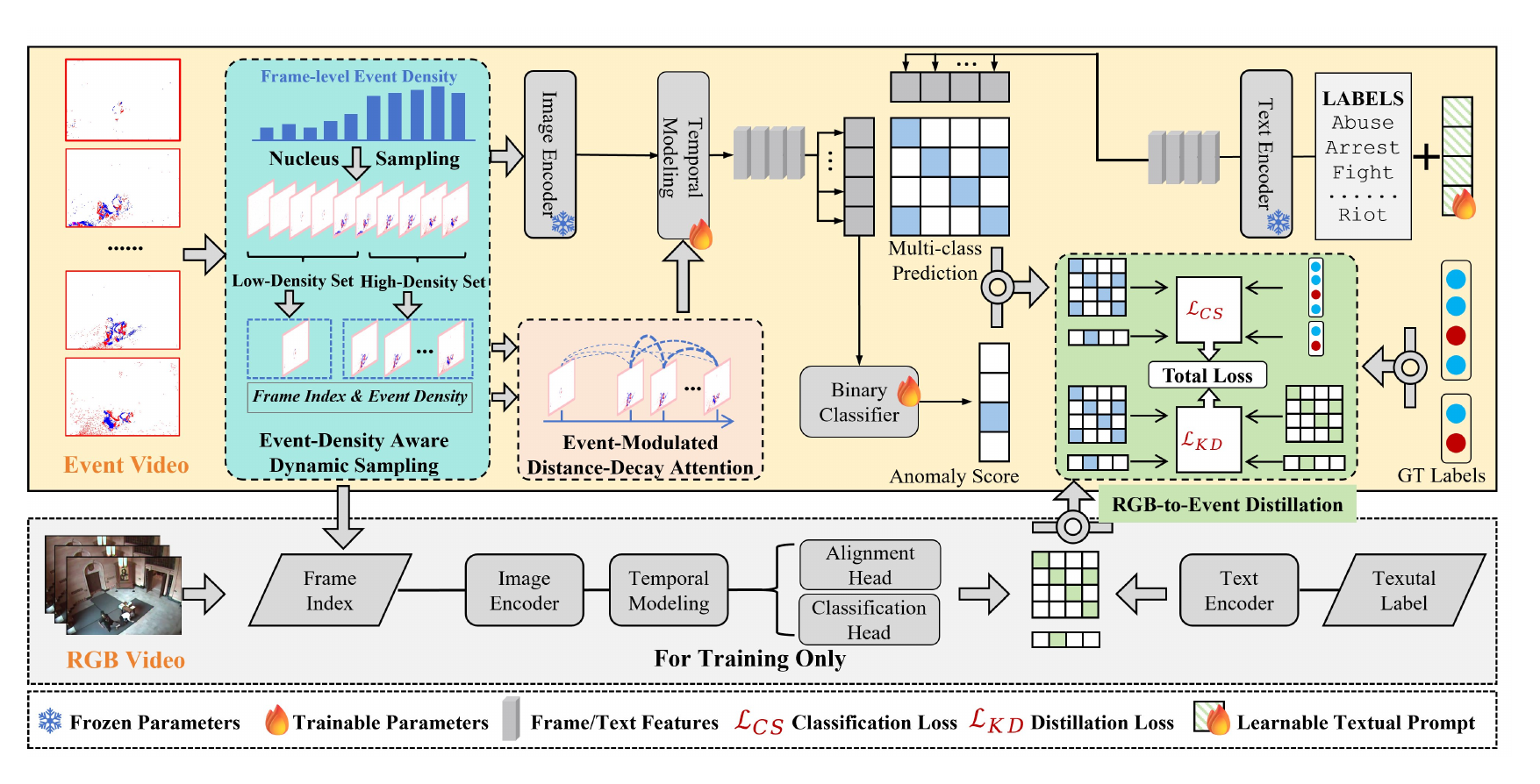}
    \caption{Overview of our proposed method EWAD.} 
    \label{fig-pipe}
\end{figure*}

The overall framework of EWAD is illustrated in Figure~\ref{fig-pipe}. Our method requires paired event and RGB modalities during the training phase but operates solely on event streams during inference. Following feature extraction, we introduce a dynamic sampling strategy EDS to prioritize temporally informative segments that are more likely to contain abnormal patterns. This allows the model to focus on frames with higher anomaly potential during training. Next, we propose a temporal modeling mechanism EDA to model long-range temporal dependencies within the sparse event sequences by jointly encoding event density and inter-event intervals. Then temporally enhanced features are forwarded to two lightweight prediction heads: a binary classification head for frame-level anomaly confidence scores, and a multi-class alignment head for semantic category prediction. In addition to conventional hard supervision via MIL, we further incorporate a soft supervision paradigm based on cross-modal distillation from an RGB-based teacher model. Specifically, we distill both the binary anomaly scores and multi-class predictions to guide the learning of the event-based student model. This multi-level distillation significantly improves the model’s discriminative capacity under weak supervision, enabling more effective anomaly detection without relying on dense annotations.
Finally, we leverage both the temporal anomaly scores and the spatial distribution of event activations to generate training-free spatial anomaly heatmaps.

\subsection{Event-Density Aware Dynamic Sampling}
Event cameras produce asynchronous, high-temporal-resolution data triggered by luminance changes, resulting in inherently sparse and non-uniform information distribution over time. Notably, anomalous activities often cause abrupt intensity variations, leading to localized bursts of event density. In this context, uniform sampling strategies may waste resources on redundant frames from low-activity regions while missing discriminative cues in dynamic segments. To address this, we propose an event-density aware dynamic sampling strategy that adaptively selects frames based on temporal event density, enabling the model to focus on regions with high potential anomaly signals. This not only enhances temporal modeling {effect} but also improves anomaly sensitivity under weak supervision.

Specifically, the event stream is partitioned into discrete frames. Let $n_i$ denote the number of events occurring within the $i^{th}$ frame. The event density $d_i$ for this frame is defined as the proportion of events in the $i^{th}$ frame relative to the total number of events across all frames, and is computed as:
\begin{equation}
d_i = \frac{n_i}{\sum_{j} n_j}
\end{equation}

Given the temporal sequence and corresponding event density information, our goal is to prioritize frames from high-density regions, where anomalies are more likely to occur, while retaining frames from low-density regions to preserve semantic context such as background information. This renders traditional top-k sampling suboptimal, as it rigidly excludes low-density frames, potentially discarding informative content. 
Inspired by sampling strategies in LLMs, we adopt a modified dual-interval nucleus sampling. We first define a density threshold $\tau_{d}$ (e.g., 0.95) and sort all event frames in descending order of density. We then accumulate densities until their sum exceeds $\tau_{d}$, designating these frames as the high-density set, with the remaining forming the low-density set. Based on a predefined sampling ratio (e.g., 8:2), we severally perform multinomial sampling without replacement within each set to select representative key frames for training. 
This dynamic sampling strategy EDS effectively emphasizes informative frames from dense event regions while ensuring the inclusion of diverse low-density frames. As a result, the model benefits from enhanced discriminative capability, improved robustness, and better generalization across varying event densities.

\subsection{Event-Modulated Distance-Decay Attention}
Event streams are inherently non-uniform in time, with varying densities reflecting different motion intensities. Meanwhile, existing studies~\cite{zhou2024unified} have shown that temporal perception varies under different event densities. Therefore, incorporating event density into temporal modeling enables the model to dynamically adjust its sense of time, leading to more accurate anomaly detection across diverse motion patterns. 
To this end, we propose an {event-modulated distance-decay attention} mechanism, which integrates both temporal distance and event density to dynamically weight temporal dependencies.

Given sampled features with corresponding timestamps $t \in \mathbb{R}^{T}$ and event densities $d \in \mathbb{R}^{T}$, we first normalize timestamps per sequence to a fixed scale. For each token pair ($i,j$), the attention weight is defined as:
\begin{equation}
w_{ij} = \frac{\exp\left(-\lambda \cdot \frac{|\tilde{t}_i - \tilde{t}j|}{d_j + \epsilon}\right)}{\sum_{k=1}^{T} \exp\left(-\lambda \cdot \frac{|\tilde{t}_i - \tilde{t}_k|}{d_k + \epsilon}\right) + \epsilon}
\end{equation}
where $\lambda > 0$ controls the decay rate, and $\epsilon$ is a small constant added for numerical stability. This formulation prioritizes closer and denser tokens, enhancing long-range temporal modeling while preserving sparsity. In this work, we seamlessly integrate this plug-and-play mechanism into the temporal modeling module of VadCLIP~\cite{wu2024vadclip}. 

\subsection{RGB-to-Event Knowledge Distillation}
Due to the limited supervision provided by hard labels and the inherent learning difficulty associated with event-based data~\cite{sun2025distilling}, we introduce an RGB-based VAD model as a teacher to guide the training of the event model. 
{Furthermore, considering that foundation models like CLIP are pre-trained extensively on massive RGB-text pairs, their semantic priors are inherently tied to the RGB modality. Direct application of CLIP to sparse and asynchronous event streams often struggles to fully unlock these pre-trained priors due to the significant modality gap. Therefore, employing an RGB-based teacher serves as a crucial bridge, transferring these high-level, RGB-centric semantics to the event-based student.}
Our cross-modal knowledge distillation strategy comprises two complementary components:

\noindent\textbf{Binary classification logit-level distillation.}
To align the anomaly confidence scores between the event model and the RGB teacher model, we minimize the mean squared error between their binary classification probabilities. Let $a^e_i \in [0,1]$ and $a^r_i \in [0,1]$ denote the predicted anomaly confidence scores of the event model and RGB teacher model, respectively, for the $i^{th}$ video frame. The binary distillation loss is defined as:
\begin{equation}
\mathcal{L}_{\mathrm{bin}} = \frac{1}{T} \sum_{i=1}^T \left( a^e_i - a^r_i \right)^2
\end{equation}

\noindent\textbf{Multi-class logit-level distillation.}
Inspired by recent advances in multi-class knowledge distillation, particularly those incorporating logit standardization~\cite{sun2024logit}, we further align the detailed categorical predictions between the two modalities to enhance the event model’s capacity for fine-grained anomaly recognition. Let ${Z}^e \in \mathbb{R}^{T \times K}$ and ${Z}^r \in \mathbb{R}^{T \times K}$ represent the logits of the event model and RGB teacher model, respectively, across $K$ categories.

We first normalize the logits, the benefit of logit standardization lies in encouraging the student model to learn the inter-class relationships predicted by the teacher, rather than attempting to precisely match the raw logits. By mitigating such discrepancies, standardization alleviates the difficulty of distillation and enhances the robustness of the student model,
\begin{equation}
\hat{{Z}} = \frac{{Z} - \mu}{\sigma + \epsilon}
\end{equation}
\begin{equation}
\mu = \frac{1}{K}\sum_{k=1}^K Z_k, \quad
\sigma = \sqrt{\frac{1}{K}\sum_{k=1}^K (Z_k - \mu)^2}
\end{equation}

We then compute the temperature-scaled softmax distributions:
\begin{equation}
p^e_{i,k} = \mathrm{softmax}\left(\frac{\hat{Z}^e_{i,k}}{\tau}\right), \quad
p^r_{i,k} = \mathrm{softmax}\left(\frac{\hat{Z}^r_{i,k}}{\tau}\right)
\end{equation}
where $\tau > 0$ is a temperature parameter. The multi-class distillation loss is calculated as the averaged Kullback–Leibler divergence, weighted by $\tau^2$:
\begin{equation}
\mathcal{L}_{\mathrm{multi}} = \tau^2 \cdot \frac{1}{T} \sum_{i=1}^T \sum_{k=1}^K p^r_{i,k} \log \frac{p^r_{i,k}}{p^e_{i,k}}
\end{equation}

The total knowledge distillation loss combines both components:
\begin{equation}
\mathcal{L}_{\mathrm{KD}} = \alpha \mathcal{L}_{\mathrm{bin}} + \beta \mathcal{L}_{\mathrm{multi}}
\end{equation}
where $\alpha$ and $\beta$ are weighting coefficients. 

It is important to note that the distillation process is applied only during training. At inference time, the event model operates independently without requiring any RGB input, ensuring {both efficiency} and practical deployment.

\subsection{Training and Inference}
During training, we optimize a composite objective loss:
\begin{equation}
\mathcal{L} = \mathcal{L}_{\mathrm{CS}} + \mathcal{L}_{\mathrm{KD}}
\end{equation}
{The classification loss $\mathcal{L}_{CS}$ follows VadCLIP~\cite{wu2024vadclip}: 
\begin{equation}
\mathcal{L}_{CS} = \mathcal{L}_{bce} + \mathcal{L}_{nce} + \mathcal{L}_{cts}
\end{equation}
}
{where $\mathcal{L}_{bce}$, $\mathcal{L}_{nce}$, $\mathcal{L}_{cts}$ are binary classification, alignment and contrastive losses, respectively.
}

During inference, temporal anomaly detection is performed by jointly leveraging the classification head and alignment head to identify anomalous frames. For each identified frame, we further analyze the corresponding event data to localize potential spatial anomaly regions. Specifically, we apply a pre-defined threshold to the event map to extract candidate regions, followed by image morphological operations to refine these regions and obtain final bounding boxes.

\section{Experiments}

\subsection{Experimental Setup}

\textbf{Evaluation metrics.} 
We evaluate model performance using the frame-level Area Under the Receiver Operating Characteristic Curve (AUC) for anomaly detection. For the spatial anomaly localization task, we report the Temporal Intersection over Union (TIoU)~\cite{liu2019exploring} to measure the accuracy of localized anomalous regions over time.

\noindent\textbf{Implementation details.} 
For classification and alignment heads, we follow the setup of VadCLIP. For training, we use the Adam optimizer with a learning rate of $2 \times 10^{-5}$. The model is trained for 10 epochs with a batch size of 128. For EDS, the default sampling length is 256. Following previous work~\cite{sun2024logit}, we set the loss weighting parameters $\alpha$ and $\beta$ as 0.1 and 9, respectively, and the temperature $\tau$ for knowledge distillation as $2$. {We extensively ablated all hyperparameters, the detailed results are provided in the supplementary material.}

\subsection{Main Experimental Results}

\noindent\textbf{Temporal anomaly detection results.} As shown in Table~\ref{tab-sota}, we compare our method with existing approaches across different feature types and model architectures, including SNN-DNN, SNN-SNN, and ViT-DNN frameworks. 

{For a strictly fair comparison under the event-centric paradigm, we adapted representative RGB-based methods (i.e., DeepMIL~\cite{sultani2018real}, DMU~\cite{zhou2023dual}, and VadCLIP~\cite{wu2024vadclip}) to take our extracted event frame features as input, training and evaluating them on our simulated event benchmarks.} 

{{Our proposed method EWAD achieves the best performance across all evaluated datasets. Specifically, on the event-camera data of UCF-Crime, EWAD achieves 72.36\% AUC, outperforming the previous state-of-the-art (MSF~\cite{qian2025ucf}, 65.01\%) by +7.35\%. Furthermore, EWAD generalizes well to our simulated benchmarks (e.g., reaching 76.55\% on simulated UCF-Crime), highlighting the practicality of simulation-derived benchmarks for scalable event-based VAD.}} 
In addition, even when using the same ViT-based features, EWAD still shows notable improvements over other ViT-DNN methods such as DMU and VadCLIP. 
These results not only highlight the superior representational power of ViT features but also demonstrate the effectiveness of our event-specific modules. It is also worth noting that no knowledge distillation is applied on CCTV-Fight and UBnormal, yet EWAD still outperforms previous methods, underscoring the strong generalization ability of our architecture and training strategy.

\begin{table}[t]
\centering
\caption{AUC comparison on different datasets.}
\label{tab-sota}
\begin{tabular}{llcccc}
\toprule
\rule{0pt}{2ex} 
\multirow{2}{*}{Category} & \multirow{2}{*}{Method} & \multirow{2}{*}{Feat-Arch.} & \multicolumn{3}{c}{AUC (\%)} \\
\cline{4-6}
 & & & UCF & CCTV & UB \\ 
\midrule
\multirow{9}{*}{\begin{tabular}[c]{@{}l@{}}\\ Event-Camera\end{tabular}} 
 & DeepMIL \cite{sultani2018real} & SNN-DNN & 55.56 & - & - \\
 & HLNet \cite{wu2020not} & SNN-DNN & 58.58 & - & - \\
 & RTFM \cite{tian2021weakly} & SNN-DNN & 52.67 & - & - \\
 & TSA \cite{joo2024clip} & SNN-DNN & 51.86 & - & - \\
\cline{2-6}
 & ResSNN \cite{fang2021deep} & SNN-SNN & 53.99 & - & - \\
 & PLIF \cite{fang2021incorporating} & SNN-SNN & 54.74 & - & - \\
 & SFormer \cite{zhou2023spikingformer} & SNN-SNN & 62.78 & - & - \\
 & MSF \cite{qian2025ucf} & SNN-SNN & 65.01 & - & - \\
 & \textbf{EWAD (Ours)} & \textbf{ViT-DNN} & \textbf{72.36} & \textbf{-} & \textbf{-} \\
\midrule
\multirow{4}{*}{Simulated} 
 & DeepMIL \cite{sultani2018real} & ViT-DNN & 61.11 & 54.98 & 56.52 \\
 & DMU \cite{zhou2023dual} & ViT-DNN & 73.47 & 57.85 & 57.46 \\
 & VadCLIP \cite{wu2024vadclip} & ViT-DNN & 73.01 & 63.08 & 53.19 \\
 & \textbf{EWAD (Ours)} & \textbf{ViT-DNN} & \textbf{76.55} & \textbf{64.17} & \textbf{58.30} \\
\bottomrule
\end{tabular}
\end{table}

\noindent\textbf{Spatial anomaly localization results.} As shown in Table~\ref{tab-tiou}, our EWAD, {leveraging a completely training-free approach} that relies solely on event modality, achieves a TIoU of 13.28\%. While the localization performance is slightly inferior to the latest RGB-based approaches, it remains competitive when compared to earlier RGB-only methods like C3D (7.20\%) and NLN (12.20\%). This demonstrates that, leveraging the low-redundant spatial representation of event data, our method can effectively deliver solid anomaly localization performance. 
{Note that the performance metrics for classic methods including TSN, C3D, and NLN are cited from Liu et al.\cite{liu2019exploring}  We include these earlier approaches to demonstrate that our {training-free}, event-based baseline can already rival or surpass established, heavily-engineered spatiotemporal RGB architectures (e.g., 3D CNNs). 
{We clarify that the spatial localization results in this work are intended as a feasibility study, aiming to explore whether event data can support spatial anomaly localization.} Meanwhile, the relatively lower TIoU compared to RGB-based methods is primarily attributed to limited temporal discriminability of event-only models.

\begin{table}[htbp]
    \centering
    \begin{minipage}[t]{0.48\textwidth}
        \centering
        \caption{TIoU comparison on UCF-Crime.}
        \label{tab-tiou}
        \begin{tabular}{lccc}
        \hline
        \multirow{1}{*}{Method} & \multirow{1}{*}{Modality} & \multicolumn{1}{c}{TIoU(\%)} \\
        \hline
        TSN \cite{wang2016temporal}     & RGB   &  2.60 \\
        C3D \cite{tran2015learning}     & RGB   &  7.20 \\
        NLN \cite{wang2018non}          & RGB   & 12.20 \\
        Liu et al. \cite{liu2019exploring} & RGB  & 16.40 \\
        VadCLIP \cite{wu2024vadclip}    & RGB   & 22.05 \\
        STPrompts \cite{wu2024weakly}   & RGB   & 23.09 \\
        \hline
        {EWAD(Ours)}                    & Event & 13.28 \\
        \hline
        \end{tabular}
    \end{minipage}
    \hfill 
    \begin{minipage}[t]{0.48\textwidth}
        \centering
        \caption{Impact of each component on UCF-Crime.} 
        \label{tab-ablation}
        \begin{tabular}{@{}ccccc@{}}
        \hline
        EDS & EDA & B-KD & M-KD & AUC (\%) \\ \hline
        ×          & ×          & ×          & ×          & 73.01 \\
        \checkmark & ×          & ×          & ×          & 73.94 \\
        \checkmark & \checkmark & ×          & ×          & 74.17 \\
        ×          & ×          & \checkmark & \checkmark & 74.45 \\
        \checkmark & \checkmark & \checkmark & ×          & 74.53 \\ 
        \checkmark & \checkmark & ×          & \checkmark & 74.58 \\ 
        \checkmark & \checkmark & \checkmark & \checkmark & 76.55 \\ 
        \hline
        \end{tabular}
    \end{minipage}
\end{table}

\subsection{Ablation Studies}
To assess the contribution of each core component, we conduct comprehensive ablation experiments on UCF-Crime dataset. Table~\ref{tab-ablation} reports the frame-level AUC under different model configurations. As we can see, each component brings clear performance gains. Specifically, EDS demonstrates its effectiveness in prioritizing informative event regions and reducing redundancy. Adding EDA further enhances temporal modeling, yielding a sustained improvement over the EDS-only setting, which validates the importance of jointly encoding event density and temporal intervals. Notably, the RGB-to-event knowledge distillation strategy alone leads to a substantial 1.44\% boost over the baseline, highlighting the advantage of transferring high-level semantic priors from the RGB domain. When all components are integrated, the model achieves the highest performance, with a 3.54\% absolute gain over the base configuration, confirming the complementary benefits of our design. The last three rows report the performance under different distillation strategies. Here, B-KD refers to binary knowledge distillation, while M-KD denotes multi-class knowledge distillation. Combining both B-KD and M-KD yields additional improvements, suggesting that binary-level and multi-class-level supervision provide complementary guidance to enhance the discriminative capability of the student model.
{
It is worth noting that our framework is designed to establish a strong uni-modal event baseline, enabling a clearer evaluation of the intrinsic potential of event data without relying on RGB inputs.
}

\subsection{Visualization }

\noindent\textbf{Visualization of temporal anomaly detection.} As shown in Figure~\ref{fig-vis}, the pink-shaded areas indicate the ground-truth anomalous segments, while the blue curves represent the anomaly scores predicted by our EWAD. Notably, certain scene transitions, such as opening or closing credits, share visual patterns with dynamic events, which introduces significant challenges in accurately detecting anomalies. Nevertheless, EWAD consistently generates high confidence scores that align with the ground-truth abnormal regions, while maintaining low responses in normal intervals, demonstrating its discriminative capability. However, compared to RGB-based methods, the detected abnormal segments appear less temporally continuous, and the distinction between normal and abnormal regions is relatively weaker, highlighting the potential for improvement through multimodal (RGB+Event) fusion. 

\noindent\textbf{Visualization of spatial anomaly localization.} Figure~\ref{fig-spatial} illustrates spatial anomaly localization results on the simulated event data. The figure shows two bounding boxes: the green box represents the ground truth (GT) anomaly region, while the pink box indicates our predicted anomaly region obtained from the event frames. As shown, the predicted bounding boxes based on event data align well with the true anomaly locations. This highlights the superiority of the event data for spatial anomaly detection.

\begin{figure}[t]
    \centering
    \includegraphics[width=0.99\linewidth]{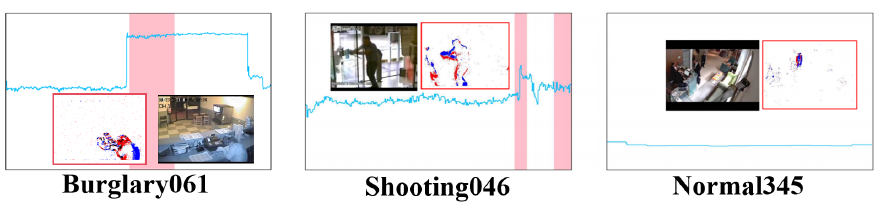}
    \caption{Qualitative results of temporal anomaly detection.} 
    \label{fig-vis}
\end{figure}

\begin{figure}[t]
    \begin{minipage}[t]{0.66\textwidth}
        \centering
        \includegraphics[width=\linewidth]{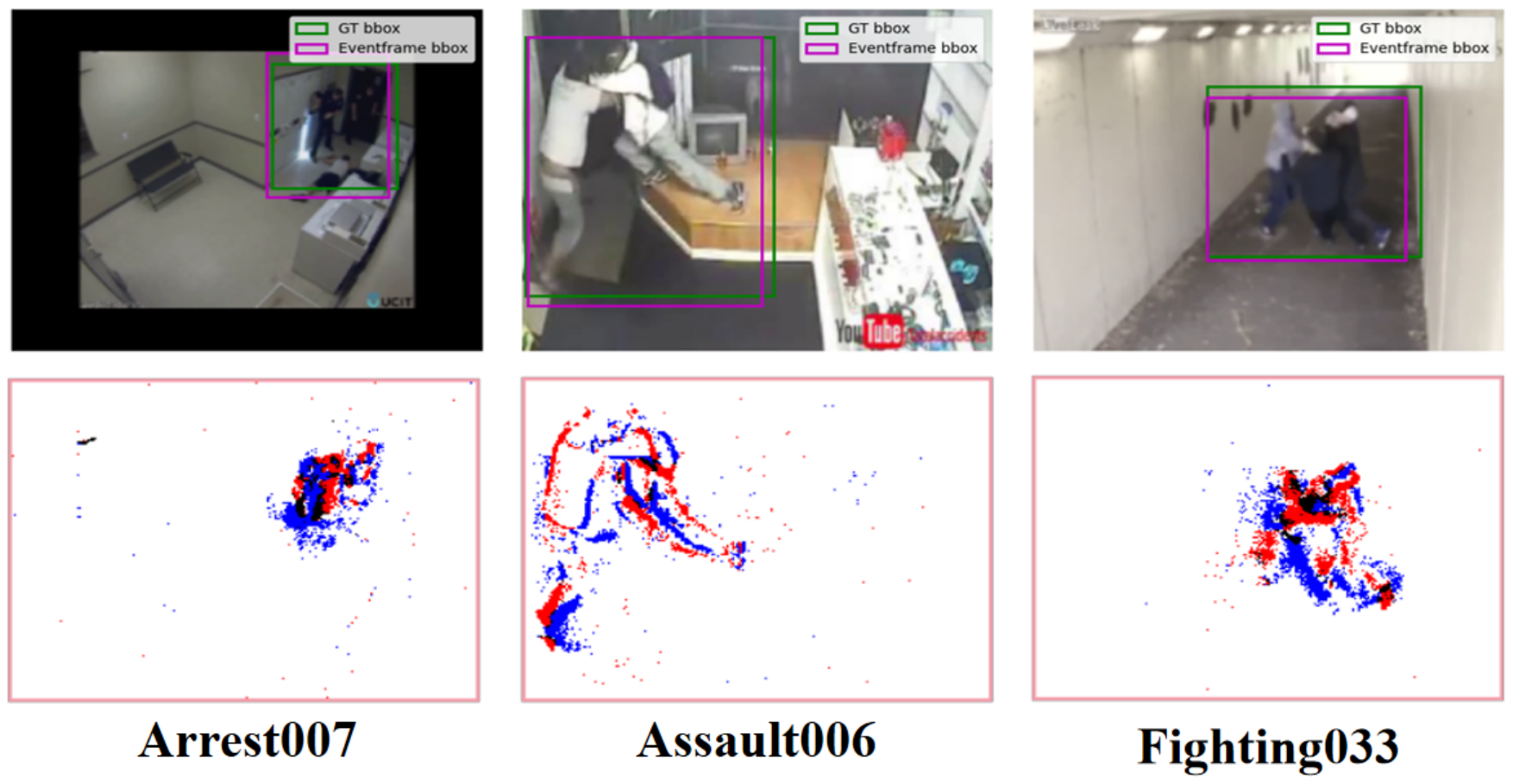}
        \caption{Visualization of spatial anomaly localization.} 
        \label{fig-spatial}
    \end{minipage}
    \hfill 
    \begin{minipage}[t]{0.32\textwidth}
        \centering
        \includegraphics[width=\linewidth]{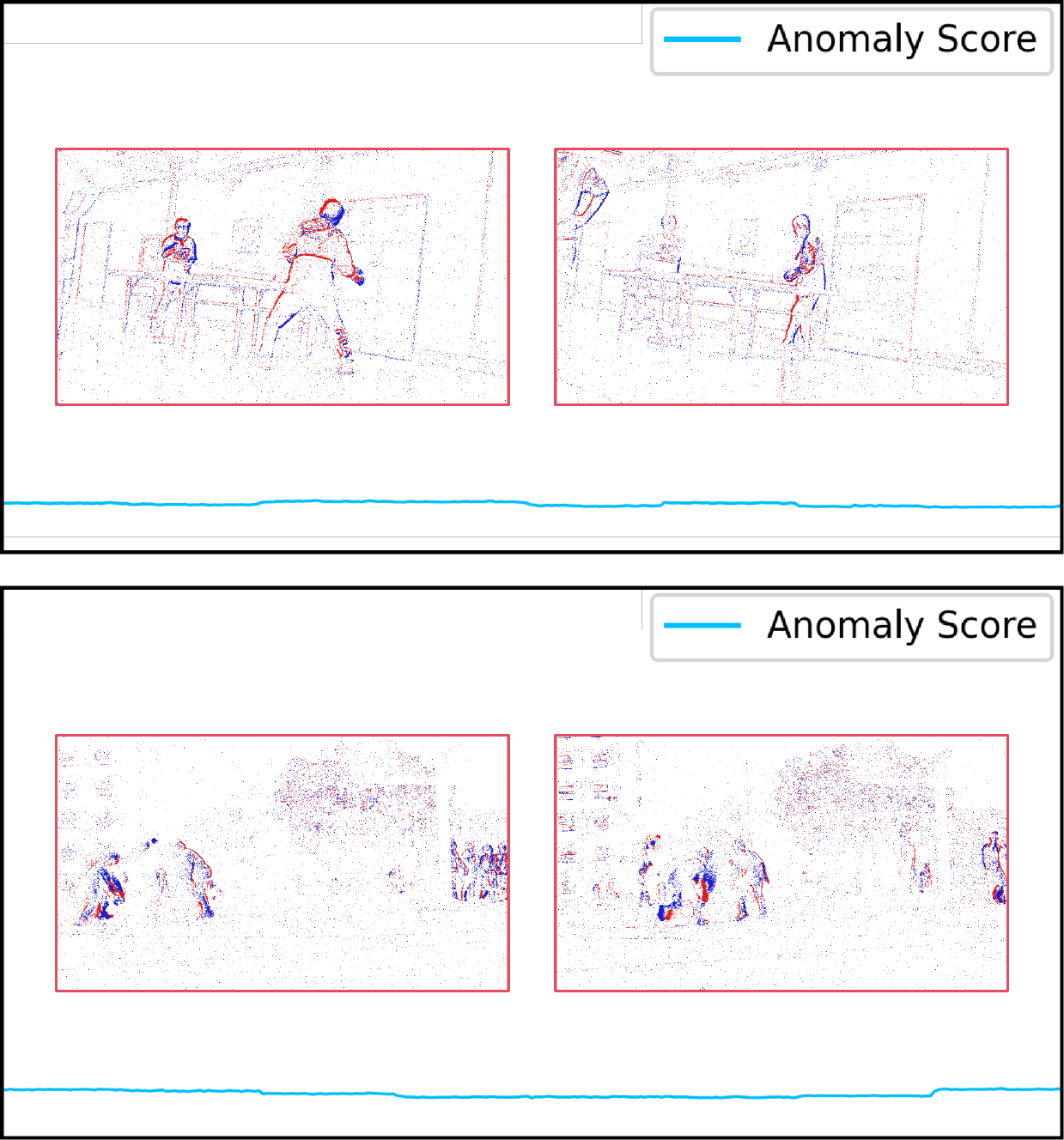}
        \caption{Anomaly scores of EWAD for examples in EventVOT.} 
        \label{fig-eventvot}
    \end{minipage}

\end{figure}

\subsection{Real Event Data Validation}

Event streams in this work are generated using the v2e simulator. As a simulation-based method, the quality of the generated events is influenced by the source RGB videos (e.g., motion blur or exposure conditions), which may introduce artifacts. Despite these limitations, the absence of large-scale event-stream datasets remains a key bottleneck for event-based VAD. Capturing rare anomalous events with real event cameras is extremely challenging in practice, making simulation-based benchmarks a practical step toward enabling systematic study in this direction. Moreover, VAD primarily relies on motion dynamics and temporal irregularities rather than fine-grained visual textures. As a result, moderate simulation artifacts are less detrimental for VAD tasks than for low-level vision problems.

To further assess the real-world applicability of our approach, we conducted validation experiments on 18 randomly selected normal video clips from the real-world EventVOT dataset~\cite{wang2024event}, covering diverse scene categories. 
Our results show that 17 out of 18 samples yield consistently low anomaly scores (representative examples in Figure~\ref{fig-eventvot}), suggesting that EWAD learns motion semantics rather than relying on artifacts introduced by simulation.

\section{Conclusion}
In this work, we take a foundational step toward event-based video anomaly detection by establishing both a strong baseline EWAD and the first set of large-scale RGB–event benchmarks. 
EWAD integrates event-density aware dynamic sampling, density-modulated temporal modeling, and cross-modal knowledge transfer into a unified framework, showing that carefully designed mechanisms can effectively bridge the gap between asynchronous event signals and high-level anomaly understanding. Beyond temporal-level detection, our exploration of spatial localization further highlights the potential of event data for fine-grained abnormality analysis. 
Comprehensive experiments validate the strength and generality of our approach across multiple datasets. Future work will explore constructing real-world event-centric VAD datasets, exploring multimodal fusion with RGB and audio, and developing foundation models tailored to event streams.
The paper ends with a conclusion. 

\clearpage  

\bibliographystyle{splncs04}
\bibliography{main}
\end{document}